\documentclass[12pt]{article}
\usepackage{amsmath,amssymb,amsthm,bm}
\usepackage{times}
\usepackage{color}
\usepackage{multirow}
\usepackage[authoryear]{natbib}
\usepackage{rotating}
\usepackage{latexsym}

\usepackage{algorithm}
\usepackage{algpseudocode}
\usepackage{setspace}

\newcommand{\captionfonts}{\normalsize}

\makeatletter
\long\def\@makecaption#1#2{%
\vskip\abovecaptionskip
\sbox\@tempboxa{{\captionfonts #1: #2}}%
\ifdim \wd\@tempboxa >\hsize
{\captionfonts #1: #2\par}
\else
\hbox to\hsize{\hfil\box\@tempboxa\hfil}%
\fi
\vskip\belowcaptionskip}
\makeatother

\newtheorem{theorem}{Theorem}

\newtheorem{corollary}{Corollary}

\begin{document}

\hspace{13.9cm}1

\ \vspace{20mm}\\

\begin{center}
{\LARGE Full-Span Log-Linear Model and Fast Learning Algorithm}
\end{center}
\ \\
{\bf \large Kazuya Takabatake, Shotaro Akaho}\\
HIIRI, AIST\\
%


\thispagestyle{empty}
\markboth{}{NC instructions}
\ \vspace{-0mm}\\
%
\begin{center} {\bf Abstract} \end{center}
The full-span log-linear(FSLL) model introduced in this paper is considered an $n$-th order Boltzmann machine, where $n$ is the number of all variables in the target system.
Let $X=(X_0,...,X_{n-1})$ be finite discrete random variables that can take $|X|=|X_0|...|X_{n-1}|$ different values.
The FSLL model has $|X|-1$ parameters and can represent arbitrary positive distributions of $X$.
The FSLL model is a ``highest-order" Boltzmann machine; nevertheless, we can compute the dual parameter of the model distribution, which plays important roles in exponential families, in $O(|X|\log|X|)$ time.
Furthermore, using properties of the dual parameters of the FSLL model, we can construct an efficient learning algorithm.
The FSLL model is limited to small probabilistic models up to $|X|\approx2^{25}$; however, in this problem domain, the FSLL model flexibly fits various true distributions underlying the training data without any hyperparameter tuning.
The experiments presented that the FSLL successfully learned six training datasets such that $|X|=2^{20}$ within one minute with a laptop PC.
\section{Introduction}\label{sec: intro}
The main purpose of this paper is as follows:
\begin{itemize}
\item To introduce the full-span log-linear(FSLL) model and a fast learning algorithm,
\item To demonstrate the performance of the FSLL model by experiments.
\end{itemize}

Boltzmann machines \citep{Ackley1985} are multivariate probabilistic models that are widely used in the field of machine learning.
Here, let us consider a Boltzmann machine with $n$ binary variables $X=(X_0,..,X_{n-1})(X_i=0,1)$.
In this paper, we handle fully connected Boltzmann machines with no hidden variables and no temperature parameter.
A Boltzmann machine represents the following distribution $p_\theta$, which we refer to as the {\it model distribution}.
\begin{multline}\label{eq: BM}
p_\theta(x)=Z(\theta)^{-1}e^{l_\theta(x)},\quad
x_i\in\{0,1\},\quad
Z(\theta)=\sum_{x\in X}e^{l_\theta(x)},\\
l_\theta(x)=\sum_{0\le i<j<n}\theta_{ij}x_ix_j+\sum_{0\le i<n}\theta_{in}x_i.
\end{multline}

In Boltzmann machines, learning is achieved by minimizing $ KL (p_d\|p_\theta)$, where $ KL (*\|*)$ is the Kullback–Leibler(KL-) divergence and $p_d$ is the empirical distribution of the training data.
One straightforward method to minimize $KL(p_d\|p_\theta)$ is to use a gradient vector whose components are
\begin{equation}\label{eq: gradBM}
\frac{\partial KL(p_d\|p_\theta)}{\partial\theta_{ij}}
=\left<X_iX_j\right>_{p_\theta}-\left<X_iX_j\right>_{p_d}
\end{equation}
\citep{Ackley1985} and to apply the gradient descent method or quasi-Newton method \citep{dennis1977quasi}.
In evaluating Eq.\eqref{eq: gradBM}, the computational cost to evaluate the term $\left<X_iX_j\right>_{p_\theta}$ is significant because we need to evaluate this term every time $\theta$ is modified.

One disadvantage of the Boltzmann machine is its insufficient ability to represent distributions.
The Boltzmann machine has only $n(n+1)/2$ parameters while the dimension of the function space spanned by the possible distributions of $X$ is $|X|-1$ ($-1$ comes from the constraint $\sum_{x\in X}p(x)=1$).

One way to reduce this disadvantage is to introduce higher-order terms into the function $l_\theta(X)$.
For example, third-order Boltzmann machines represent the following distribution $p_\theta$ \citep{Sejnowski1986}:
\begin{multline*}
p_\theta(x)=Z(\theta)^{-1}e^{l_\theta(x)},\quad
x_i\in\{0,1\},\quad
Z(\theta)=\sum_{x\in X}e^{l_\theta(x)},\\
l_\theta(x)=\sum_{0\le i<j<k<n}\theta_{ijk}x_ix_jx_k
+\sum_{0\le i<j<n}\theta_{ijn}x_ix_j+\sum_{0\le i<n}\theta_{inn}x_i.
\end{multline*}
Here, ``order'' means the number of variables on that a function depends.
This definition of order is not limited to binary variables.
For example, if $f(x_0,x_1,x_2)$ ignores $x_2$, that is, $x_2$ does not affect the value $f(x_0,x_1,x_2)$, the order of $f$ is two.
The $k$-th-order Boltzmann machine has up to $k$-th-order terms in $l_\theta$.
Since the $n$-th-order Boltzmann machine has arbitrary order of terms, it can represent arbitrary positive distributions\footnote{\normalsize distributions such that $\forall x, p(x)>0$} of $X$.

However, introducing higher-order terms leads to an enormous increase in computational cost because the $k$-th order Boltzmann machine has $\sum_{i=1}^k\binom ni\approx n^k/k$ parameters.

The FSLL model introduced in this paper can be considered an $n$-th order Boltzmann machine\footnote{\normalsize Furthermore, the FSLL model is not limited to binary variables.}, where $n$ is the number of all variables in the target system.
The FSLL model has $|X|-1$ parameters and can represent arbitrary positive distributions.
Since the FSLL model is a ``highest-order'' Boltzmann machine, the learning of FSLL is expected to be very slow.
However, we propose a fast learning algorithm.
For example, this algorithm can learn a joint distribution of 20 binary variables within 1 minute with a laptop PC.

Since the FSLL model has full degrees of freedom, a regularization mechanism to avoid overfitting is essential.
For this purpose, we used a regularization mechanism based on the minimum description length principle \citep[Chapter 8]{Rissanen07}.

The remainder of this paper is organized as follows.
In Section 2, we present the FSLL model and its fast learning algorithm.
In Section 3, we demonstrate the performance of the FSLL model by experiment.
In Section 4, we discuss the advantages/disadvantages of the FSLL model.
In Section 5, we present the conclusions and extensions of the paper.

\section{Full-Span Log-Linear Model}\label{sec: fsll}
Before introducing the FSLL model, we define the notations used in this paper.
A random variable is denoted by a capital letter, such as $X$, and the value that $X$ takes is indicated by a lower case letter, such as $x$.
$X$ also denotes the set of values that the variable $X$ can take; thus, $|X|$ denotes the number of values that $X$ can take.
$\left<f\right>_p$ denotes the expectation of $f(X)$ with distribution $p(X)$, that is, $\left<f\right>_p=\sum_{x\in X}p(x)f(x)$.
The differential operator $\partial/\partial\theta_y$ is abbreviated as $\partial_y$.

\subsection{Model Distribution}
The FSLL model is a multivariate probabilistic model designed for a system that has $n$ discrete finite variables $X_0,...,X_{n-1}$, where $X_i$ takes an integer value in $[0,|X_i|)$.
The FSLL model has parameters $\theta=\{\theta_y\}$, where $y=(y_0,...,y_{n-1})$ is a vector such that $y_i\in X_i$.
The model distribution of the FSLL model is the following $p_\theta$:
\begin{multline}\label{eq: fsllmodel}
p_\theta(x)=Z(\theta)^{-1}e^{l_\theta(x)},\quad
x_i\in X_i,\quad
Z(\theta)=\sum_{x\in X}e^{l_\theta(x)},\\
l_\theta(x)=\sum_{y\in X}\theta_y\Phi_y(x),\quad
\Phi_y(x)=\prod_{i=0}^{n-1}\phi^i_{y_i}(x_i).
\end{multline}
In Eq.\eqref{eq: fsllmodel}, $\{\phi^i_{y_i}\}(y_i\in X_i)$ are $|X_i|$ linearly independent functions of $X_i$, which we refer to as the {\it local basis functions}, and $\{\Phi_y\}(y\in X)$ are $|X|$ functions of $X=(X_0,...,X_{n-1})$, which we refer to as the {\it(global) basis functions}.
Using the following theorem recursively, we can prove that the global basis functions are linearly independent functions of $X$.
\begin{theorem}\label{thm: independence}
If $\{f_i\}(i\in I)$ are linearly independent functions of $X_0$, and $\{g_j\}(j\in J)$ are linearly independent functions of $X_1$, then $\{f_i(X_0)g_j(X_1)\}(i\in I,j\in J)$ are linearly independent functions of $(X_0,X_1)$.
\end{theorem}
The proof is provided in the Appendix.

In the FSLL model, we determine the local basis functions as follows:
\begin{description}
\item[Case $|X_i|=2^k$:]
\begin{align}
&H_1=(1),\quad
H_{2^{k+1}}=\begin{pmatrix}
H_{2^k}&H_{2^k}\\
H_{2^k}&-H_{2^k}
\end{pmatrix},\label{eq: WHmatrix}\\
&\begin{pmatrix}
\phi^i_0(0)&...&\phi^i_0(|X_i|-1)\\
\vdots&&\vdots\\
\phi^i_{|X_i|-1}(0)&...&\phi^i_{|X_i|-1}(|X_i|-1)
\end{pmatrix}
=H_{2^k}\notag
\end{align}
$H_{2^k}$ above is referred to as the {\it Walsh--Hadamard matrix} \citep{Pratt1969}.
\item[Else:]
\begin{equation}\label{eq: nonWH}
\phi^i_0\equiv1,\quad
\phi^i_j(l)=\begin{cases}
1&(0<j=l)\\
-1&(0<j\ne l)
\end{cases}
\end{equation}
\end{description}

Since $\Phi_0\equiv1$, an arbitrary $\theta_0$ gives the same model distribution.
Therefore, we determine as $\theta_0\equiv0$.

\subsection{Learning Algorithm}\label{sec: learning algorithm}
\begin{algorithm}[h]
\caption{Learning algorithm}\label{alg: learn}
$\theta$: parameters\\
$p_\theta$: model distribution\\
$p_d$: empirical distribution of training data\\
$\bar\theta$: dual parameters of $\theta$\\
$\epsilon$: threshold to halt the loop\\
$r_y$: regularization term for $y$ such that $\theta_y\ne0$
\begin{algorithmic}[1]
\Function {learn}{training data}
\State Compute $\bar d$ from $p_d$ (Section \ref{sec: fast algorithm})
\State Compute $r_y$(Eq.\eqref{eq: cost}) for all $y\in X$
\State\label{line: initialize} $\theta\leftarrow0$, $p_\theta\leftarrow$uniform distribution
\Loop\label{line: loop}
\State Compute $\bar\theta$ from $p_\theta$ (Section \ref{sec: fast algorithm})
\State\label{line: evaluate}Evaluate all candidates $\theta'$ and find
$\theta^1=\arg\min_{\theta'}cost(\theta')$\
\If{$cost(\theta)-cost(\theta^1)<\epsilon$}
\Return $\theta,p_\theta$
\EndIf
\State\label{line: update theta} $\theta\leftarrow\theta^1$
\State\label{line: update p_theta}Update $p_\theta$
\EndLoop
\EndFunction
\end{algorithmic}
\end{algorithm}

Algorithm \ref{alg: learn} presents the outline of the learning algorithm of the FSLL model.
This algorithm is a greedy search to find a local minimum point of $cost(\theta)$.
$cost(\theta)$ monotonically decreases as the iteration progresses.

In line \ref{line: evaluate}, {\it candidate} denotes $\theta'$ derived from $\theta$ by applying one of the following modifications on the $y$-th component:
\begin{description}
\item[Candidate derived by appending $\theta_y$:] If $\theta_y=0$, then let $\theta'_y=\arg\min_{\theta_y}cost(\theta)$.
\item[Candidate derived by adjusting $\theta_y$:] If $\theta_y\ne0$, then let $\theta'_y=\arg\min_{\theta_y}cost(\theta)$.
\item[Candidate derived by removing $\theta_y$:] If $\theta_y\ne0$, then let $\theta'_y=0$.
\end{description}

\subsubsection{Cost Function}
We use a cost function based on the minimum description length principle \citep[Chapter 8]{Rissanen07}.
Suppose that we send the training data to a receiver by transmitting the parameters and compressed data.
Since $\theta$ is a sparse vector, we transmit only indexes $y$, such that $\theta_y\ne0$, and the value of $\theta_y$.
Moreover, the index $y=(y_0,...,y_{n-1})$ is a sparse vector because higher-order basis functions are rarely used in the model distribution due to their expensive cost.
Therefore, we transmit only indexes $i$, such that $y_i\ne0$, and the values of $y_i\in[1,|X_i|)$ to transmit the sparse vector $y$.
Then, the description length\footnote{\normalsize We use ``nat" as the description length unit; thus, we use $\ln$ instead of $\log$.} to transmit the sparse vector $y$ becomes
$$\sum_{i:y_i\ne0}\underbrace{\ln n}_\text{to send $i\in[0,n)$}
+\underbrace{\ln(|X_i|-1)}_\text{to send $y_i\in[1,|X_i|)$}=\sum_{i:y_i\ne0}\ln n(|X_i|-1),$$
and the description length to transmit all index vectors $y$, such that $\theta_y\ne0$, becomes
$$\sum_{y:\theta_y\ne0}\sum_{i:y_i\ne0}\ln n(|X_i|-1).$$
The minimum description length to transmit $k$ parameters and the compressed data is estimated as \citep[Chapter 8]{Rissanen07}
$$-N\left<\ln p_\theta\right>_{p_d}+\frac k2\ln N,$$
where $k$ is the number of non-zero parameters and $N$ is the number of samples in the training data.
The total description length to transmit $y,\theta_y$ and the compressed data becomes
\begin{multline}\label{eq: MDL}
-N\left<\ln p_\theta\right>_{p_d}+\frac k2\ln N
+\sum_{y:\theta_y\ne0}\sum_{i:y_i\ne0}\ln n(|X_i|-1)\\
=-N\left<\ln p_\theta\right>_{p_d}
+\sum_{y:\theta_y\ne0}\left(\frac{\ln N}2+\sum_{i:y_i\ne0}\ln n(|X_i|-1)\right).
\end{multline}
We divide Eq.\eqref{eq: MDL} by $N$ and add $\left<\ln p_d\right>_{p_d}$ to create information geometric quantity, and obtain the following cost function:
\begin{multline}\label{eq: cost}
cost(\theta,N)=KL(p_d\|p_\theta)+r(\theta,N),\quad
r(\theta,N)=\sum_{y:\theta_y\ne0}r_y(N),\\
r_y(N)=\frac1N\left(\frac{\ln N}2+\sum_{i:y_i\ne0}\ln n(|X_i|-1)\right).
\end{multline}

\subsubsection{Fast Algorithm to Compute $\bar\theta$}\label{sec: fast algorithm}
The following vector $\bar\theta$ is referred to as the {\it dual parameter} of $\theta$, which plays important roles in the exponential families \citep[Chapter 2]{amari2016information}:
\begin{align}\label{eq: barTheta}
\bar\theta_y(\theta)&=\partial_y\ln Z(\theta)\notag\\
&=\left<\Phi_y\right>_{p_\theta}\quad
\text{(see Appendix for derivation).}
\end{align}

We identified an algorithm to compute $\bar\theta$ from $p_\theta$ in $O(|X|\log|X|)$ time.
This algorithm borrowed ideas from the multidimensional discrete Fourier transform (DFT) \citep[Chapter 7]{Smith2010}.
Here, let us consider a two-dimensional(2D-)DFT\footnote{\normalsize Not 2D-FFT but 2D-DFT.} for $|X_0|\times|X_1|$ pixels of data.
The 2D-DFT transforms $f$ into $F$ by the following equation:
\begin{multline*}
F(y_0,y_1)=\sum_{x_0,x_1}f(x_0,x_1)\Phi_{y_0y_1}(x_0,x_1),\\
\Phi_{y_0y_1}(x_0,x_1)=\exp\left(\frac{2\pi i}{|X_0|}y_0x_0\right)\exp\left(\frac{2\pi i}{|X_1|}y_1x_1\right).
\end{multline*}
To derive the value of $F(y_0,y_1)$ for a specific $(y_0,y_1)$, $O(|X_0||X_1|)$ time is required, therefore, it appears that $O(|X_0|^2|X_1|^2)$ time is required to derive $F(y_0,y_1)$ for all $y_0,y_1$.
However, 2D-DFT is usually realized in the following dimension-by-dimension manner:
\begin{equation}\label{eq: 2dDFT}
F(y_0,y_1)
=\underbrace{\sum_{x_1}\underbrace{\left(\sum_{x_0}f(x_0,x_1)\exp\left(\frac{2\pi i}{|X_0|}y_0x_0\right)\right)}_{\text{DFT by }X_0}
\exp\left(\frac{2\pi i}{|X_1|}y_1x_1\right)}
_{\text{DFT by }X_1}.
\end{equation}
In Eq.\eqref{eq: 2dDFT}, the DFT by $X_0$ for all $y_0\in X_0,x_1\in X_1$ requires $O(|X_0|^2|X_1|)$ time, and the DFT by $X_1$ for all $y_0\in X_0,y_1\in X_1$ requires $O(|X_0||X_1|^2)$ time.
Therefore, the entire 2D DFT requires $O(|X_0||X_1|(|X_0|+|X_1|))$ time that is smaller than $O(|X_0|^2|X_1|^2)$.
The key here is that the basis function $\Phi_{y_0y_1}(x_0,x_1)$ is a product of two univariate functions as follows:
\begin{multline*}
$$\Phi_{y_0y_1}(x_0,x_1)=\phi^0_{y_0}(x_0)\phi^1_{y_1}(x_1),\\
\phi^0_{y_0}(x_0)=\exp\left(\frac{2\pi i}{|X_0|}y_0x_0\right),\quad
\phi^1_{y_1}(x_1)=\exp\left(\frac{2\pi i}{|X_1|}y_1x_1\right).
\end{multline*}
We apply this principle to compute $\bar\theta$.

Here, we consider how to derive $\bar\theta$ from $p_\theta$ by the following equation:
\begin{align}\label{eq: gn}
\left<\Phi_y\right>_{p_\theta}&=\sum_x p_\theta(x)\Phi_y(x)\notag\\
&=\sum_x p_\theta(x)\prod_i\phi^i_{y_i}(x_i)\notag\\
&=\sum_{x_{n-1}}\left(...\left(\sum_{x_1}\left(\sum_{x_0}p_\theta(x)\phi^0_{y_0}(x_0)\right)\phi^1_{y_1}(x_1)\right)...\right)\phi^{n-1}_{y_{n-1}}(x_{n-1}).
\end{align}
We evaluate the right side of Eq.\eqref{eq: gn} from the innermost parenthesis to the outermost parenthesis; that is, we determine the following function, $g^i:X\to\mathbb R$ by the following recurrence sequence:
\begin{align}\label{eq: gi2gi1}
g^0(x)&=p_\theta(x),\notag\\
g^{i+1}(y_0,...,y_{i-1},y_i,x_{i+1},...,x_{n-1})
&=\sum_{x_i}g^i(y_0,...,y_{i-1},x_i,x_{i+1},...,x_{n-1})\phi^i_{y_i}(x_i).
\end{align}
Then, the equation $\bar\theta_y=g^n(y)$ holds.
Since $O(|X||X_i|)$ time is required to obtain $g^{i+1}$ from $g^i$, we can obtain $g^n$ from $g^0$ in $O(|X|\sum_i|X_i|)$ time.
In the case where $|X_i|=k$, the computational cost to derive $g^n$ from $g^0$ becomes $O(|X|kn)$.
Moreover, considering $k$ as a constant, we obtain the computational cost as $O(|X|kn)=O(|X|k\log_k|X|)=O(|X|\log|X|)$.

We can use the same algorithm to obtain the vector $\bar d$ from $p_d$.
In this case, let $g^0=p_d$.

\subsubsection{Acceleration by Walsh-Hadamard Transform}
In cases where $|X_i|$ is large, we can accelerate the computation of $\bar\theta$ by using Walsh-Hadamard transform(WHT)\citep{fino1976unified}.

Let us recall the 2D-DFT in Section \ref{sec: fast algorithm}.
If $|X_0|,|X_1|$ are powers of two, we can use the fast Fourier transform(FFT)\citep[Chapter 3]{Smith2010} for DFT by $X_0$ and DFT by $X_1$ in Eq.\eqref{eq: 2dDFT} and can reduce the computational cost to $O(|X_0||X_1|\log|X_0||X_1|)$.
We can apply this principle to computing $\bar\theta$.

Here, let us fix the values of $y_0,...,y_{i-1},x_{i+1},...,x_{|X_i|-1}$(only $i$-th component is omitted) in Eq.\eqref{eq: gi2gi1}.
Then, we obtain:
$$g^{i+1}(...,\underset i{y_i},...)
=\sum_{x_i}g^i(...,\underset i{x_i},...)\phi^i_{y_i}(x_i).$$
Using matrix notation, we obtain:
\begin{multline}\label{eq: local transform}
\begin{pmatrix}
g^{i+1}(...,\underset i0,...)\\
\vdots\\
g^{i+1}(...,\underset i{|X_i|-1},...)
\end{pmatrix}\\
=\begin{pmatrix}
\phi^i_0(0)&...&\phi^i_0(|X_i|-1)\\
\vdots&&\vdots\\
\phi^i_{|X_i|-1}(0)&...&\phi^i_{|X_i|-1}(|X_i|-1)
\end{pmatrix}
\begin{pmatrix}
g^i(...,\underset i0,...)\\
\vdots\\
g^i(...,\underset i{|X_i|-1},...)
\end{pmatrix}.
\end{multline}
We refer to this transform $\mathbb R^{|X_i|}\to\mathbb R^{|X_i|}$ as the {\it local transform}.
The local transform usually requires $O(|X_i|^2)$ time; however, if the matrix in Eq.\eqref{eq: local transform} is a Walsh-Hadamard matrix(Eq.\eqref{eq: WHmatrix}), using Walsh-Hadamard transform(WHT)\citep{fino1976unified}, we can perform the local transform in $O(|X_i|\log|X_i|)$ time\footnote{\normalsize For $k\le4$, directly multiplying the Hadamard matrix is faster than the WHT in our environment; therefore we use the WHT only in cases where $k\ge5$.}.

Since the number of all combination of $y_0,...,y_{i-1},x_{i+1},...,x_{n-1}$($i$ th component is omitted) is $|X|/|X_i|$, we can obtain the function $g^{i+1}$ from $g^i$ in $O(|X|\log|X_i|)$ time.
Moreover, $g^n$ is obtained from $g^0$ in $O(\sum_i|X|\log|X_i|)=O(|X|\log|X|)$ time.

\subsubsection{$O(1)$ Algorithm to Evaluate Candidate}
In line \ref{line: evaluate} of Algorithm \ref{alg: learn}, all three types of candidates---appending, adjusting, removing---are evaluated for all $y\in X$.
Here, we consider the following equation:
\begin{align}
\partial_y\bar\theta_y
&=\left<\Phi_y^2\right>_{p_\theta}-\bar\theta_y^2\quad
\text{(see Appendix for derivation)}\label{eq: partial_y bar_y}\\
&=1-\bar\theta^2_y\quad
(\because \Phi_y^2\equiv1).\label{eq: ordinary diff}
\end{align}
considering $\bar\theta_y$ as a univariate function of $\theta_y$ and considering Eq.\eqref{eq: ordinary diff} as an ordinary differential equation, we obtain the following general solution:
\begin{equation}\label{eq: general solution}
\bar\theta_y(\theta_y)=\tanh(\theta_y-c)\quad
\text{(see Appendix for derivation),}
\end{equation}
where $c$ is a constant determined by a boundary condition.
For example, if
$\bar\theta_y(\theta_y^0)=\bar\theta_y^0$, then $c$ is given by $c=\theta_y^0-\tanh^{-1}\bar\theta_y^0$ and Eq.\eqref{eq: general solution} becomes
\begin{equation}\label{eq: theta2barTheta}
\bar\theta_y(\theta_y)=\tanh(\theta_y-\theta_y^0+\tanh^{-1}\bar\theta_y^0).
\end{equation}
Here, let us define the follwing line $A_y(\theta^0)$ in $P$:
\begin{equation}\label{eq: def A_y}
A_y(\theta^0)=\{\theta|\forall y'\ne y,\ \theta_y=\theta^0_y\}.
\end{equation}
Equation\eqref{eq: theta2barTheta} demonstrates that if $\bar\theta_y^0$ is known, then we can derive any $\bar\theta_y$ at a point on the line
$A_y(\theta^0)$ in $O(1)$ time.

Here, the gradient vector of $KL(p_d\|p_\theta)$ is given by the following equation:
\begin{equation}\label{eq: gradKL}
\partial_y KL(p_d\|p_\theta)=\bar\theta_y-\bar d_y,\quad
\bar d_y=\left<\Phi_y\right>_{p_d}\quad
\text{(see Appendix for derivation).}
\end{equation}
Therefore, for $\theta\in A_y(\theta^0)$, we can obtain $KL(p_d\|p_\theta)-KL(p_d\|p_{\theta^0})$ by integrating Eq.\eqref{eq: gradKL} as follows:
\begin{align}\label{eq: KL(p_d||p_theta)}
KL(p_d\|p_\theta)-KL(p_d\|p_{\theta^0})
&=\int_{\theta_y^0}^{\theta_y}\bar\theta_y(u)-\bar d_y\ du\notag\\
&=\frac{1+\bar d_y}2\ln\frac{1+\bar\theta^0_y}{1+\bar\theta_y}
+\frac{1-\bar d_y}2\ln\frac{1-\bar\theta^0_y}{1-\bar\theta_y}\\
&\text{(see Appendix for derivation)}.\notag
\end{align}
Then, we can evaluate $\Delta(\theta')=cost(\theta')-cost(\theta^0)$ for three types of candidates $\theta'$ as follows:
\begin{description}
\item[Appending $\theta_y$:]By Eq.\eqref{eq: gradKL}, $KL(p_d\|p_\theta)$ is minimized when $\bar\theta_y=\bar d_y$.
Therefore,
\begin{equation}\label{eq: candidate by appending}
\Delta=\frac{1+\bar d_y}2\ln\frac{1+\bar\theta^0_y}{1+\bar d_y}
+\frac{1-\bar d_y}2\ln\frac{1-\bar\theta^0_y}{1-\bar d_y}+r_y.
\end{equation}
\item[Adjusting $\theta_y$:]By Eq.\eqref{eq: gradKL}, $KL(p_d\|p_\theta)$ is minimized when $\bar\theta_y=\bar d_y$.
Therefore,
$$\Delta=\frac{1+\bar d_y}2\ln\frac{1+\bar\theta^0_y}{1+\bar d_y}
+\frac{1-\bar d_y}2\ln\frac{1-\bar\theta^0_y}{1-\bar d_y}.$$
\item[Removing $\theta_y$:]$\theta_y$ becomes 0.
Therefore,
$$\Delta=\frac{1+\bar d_y}2\ln\frac{1+\bar\theta^0_y}{1+\bar\theta_y(0)}
+\frac{1-\bar d_y}2\ln\frac{1-\bar\theta^0_y}{1-\bar\theta_y(0)}-r_y.$$
\end{description}

\paragraph{Reducing Computational Cost of Evaluating Candidates}
Among the three types of candidates---appending, adjusting, removing---the group of candidates by appending $\theta_y$ has almost $|X|$ candidates because $\theta_y$ is a very sparse vector.
Therefore, it is important to reduce the computational cost to evaluate candidates by appending.
Evaluating $\Delta$ in Eq.\eqref{eq: candidate by appending} is an expensive task for a central processing unit(CPU) because it involves logarithm computation\footnote{\normalsize Logarithm computation is 30 times slower than addition or multiplication in our environment.}.

$\Delta$ in Eq.\eqref{eq: candidate by appending} has the following lower bound $\underline\Delta$:
\begin{align}
\Delta&\ge\underline\Delta\notag\\
&=-\frac{(\bar\theta^0_y-\bar d_y)^2}{1-(\bar\theta^0_y)^2}+r_y\quad
\text{(see Appendix for derivation)}\label{eq: lower bound}.
\end{align}
Evaluating $\underline\Delta$ is much faster than evaluating $\Delta$.
In the candidate evaluation, the candidate having lower cost is the ``winner''.
If a candidate's lower bound is greater than the champion's cost---the lowest cost ever found---the candidate has no chance to win; therefore, we can discard the candidate without precise evaluation of $\Delta$.

\begin{algorithm}[h]
\caption{Details of line \ref{line: evaluate} of Algorithm \ref{alg: learn}}
\label{alg: evaluate}
\begin{algorithmic}[1]
\State $\Delta_{\theta^1}\leftarrow0$
\ForAll{$y\in X$}
\If{$\theta_y=0$}
\State $\theta'\leftarrow$candidate by appending $\theta_y$
\If{$\underline\Delta(\theta')<\Delta_{\theta^1}$}\label{line: skip evaluation}
\If{$\Delta(\theta')<\Delta_{\theta^1}$}
$\ \theta^1_y\leftarrow\theta'_y$,
$\Delta_{\theta^1}\leftarrow\Delta(\theta')$,
$y^1\leftarrow y$
\EndIf
\EndIf
\Else
\State $\theta'\leftarrow$candidate by adjusting $\theta_y$
\If{$\Delta(\theta')<\Delta_{\theta^1}$}
$\ \theta^1_y\leftarrow\theta'_y$,
$\Delta_{\theta^1}\leftarrow\Delta(\theta')$,
$y^1\leftarrow y$
\EndIf
\State $\theta'\leftarrow$candidate by removing $\theta_y$
\If{$\Delta(\theta')<\Delta_{\theta^1}$}
$\ \theta^1_y\leftarrow\theta'_y$,
$\Delta_{\theta^1}\leftarrow\Delta(\theta')$,
$y^1\leftarrow y$
\EndIf
\EndIf
\EndFor
\end{algorithmic}
\end{algorithm}

Algorithm \ref{alg: evaluate} presents the details of line \ref{line: evaluate} of Algorithm \ref{alg: learn}.
In line \ref{line: skip evaluation}, if $\underline\Delta(\theta'_y)\ge\Delta_{\theta^1}$, then the candidate $\theta'$ is discarded, and the evaluation of $\Delta(\theta'_y)$ is skipped.
This skipping effectively reduces the computational cost of evaluating candidates\footnote{\normalsize In our environment, this skipping makes the evaluation of candidates more than ten times faster.}.

\subsubsection{Updating $\theta$ and $p_\theta$}
Let $\theta^0$ denote $\theta$ before updating and $\theta^1$ denote $\theta$ after updating in line \ref{line: update theta} of Algorithm \ref{alg: learn}.
The $\theta^1$ differs from $\theta^0$ only at $y'$-th component.
Therefore,
$$l_{\theta^1}(x)=l_{\theta^0}(x)+(\theta^1_{y'}-\theta^0_{y'})\Phi_{y'}(x),$$
and
\begin{align}
p_{\theta^1}(x)&=Z(\theta^1)^{-1}\exp E_{\theta^1}(x)\notag\\
&=Z(\theta^1)^{-1}\exp\left(E_{\theta^0}(x)+(\theta^1_{y'}-\theta^0_{y'})\Phi_{y'}(x)\right)
\notag\\
&=\frac{Z(\theta^0)}{Z(\theta^1)}
p_{\theta^0}(x)\exp\left((\theta^1_{y'}-\theta^0_{y'})\Phi_{y'}(x)\right)
\label{eq: update theta}.
\end{align}
Summing Eq.\eqref{eq: update theta} for all $x\in X$, we obtain
$$1=\frac{Z(\theta^0)}{Z(\theta^1)}
\sum_x p_{\theta^0}(x)
\exp\left((\theta^1_{y'}-\theta^0_{y'})\Phi_{y'}(x)\right).$$
Therefore,
\begin{multline}\label{eq: update p_theta}
p_{\theta^1}(x)=\frac{Z(\theta^0)}{Z(\theta^1)}
p_{\theta^0}(x)\exp\left((\theta^1_{y'}-\theta^0_{y'})\Phi_{y'}(x)\right),\\
\frac{Z(\theta^0)}{Z(\theta^1)}
=\left(\sum_x p_{\theta^0}(x)\exp\left((\theta^1_{y'}-\theta^0_{y'})\Phi_{y'}(x)\right)\right)^{-1}.
\end{multline}

\begin{algorithm}[h]
\caption{Details of lines \ref{line: update theta}--\ref{line: update p_theta} of Algorithm \ref{alg: learn}}
\label{alg: update}
\begin{algorithmic}[1]
\State $sum\leftarrow0,\ c_+\leftarrow \exp(\theta^1_{y'}-\theta_{y'}),\ c_-\leftarrow 1/c_+$
\State $\theta_{y^1}\leftarrow\theta^1 _{y^1}$
\ForAll{$x\in X$}
\If{$\Phi_{y^1}(x)=1$} $p_\theta(x)\leftarrow c_+ p_\theta(x)$
\Else $\ p_\theta(x)\leftarrow c_- p_\theta(x)$
\EndIf
\State $sum\leftarrow sum+p_\theta(x)$
\EndFor
\ForAll{$x\in X$}$\ p_\theta(x)\leftarrow p_\theta(x)/sum$
\EndFor
\end{algorithmic}
\end{algorithm}

Algorithm \ref{alg: update} presents the details of lines \ref{line: update theta}--\ref{line: update p_theta} in Algorithm \ref{alg: learn}.
This algorithm requires $O(|X|)$ time.
It should be noted that the expensive exponential computation
$\exp\left((\theta^1_{y'}-\theta^0_{y'})\Phi_{y'}(x)\right)$ is not used in the for-loop.

\subsubsection{Memory Requirements}
In the FSLL model, most of the memory consumption is dominated by four large tables for $p_\theta$, $\bar\theta$, $\bar d$, and $r$, and each stores $|X|$ floating point numbers.
On the other hand, $\theta$ does not require a large amount of memory because it is a sparse vector.

For example, if the FSLL model is allowed to use 4 GB of memory, it can handle up to 26 binary variables, 16 three-valued variables($|X_i|=3$), and 13 four-valued variables.

\subsection{Convergence to target distribution}
One major interest about Algorithm \ref{alg: learn} in the previous subsection is whether the model distribution converges to a target distribution at the limit of $N\to\infty$ or not, where $N$ is the number of samples in the training data.
As a result, we can guarantee this convergence.

Let us define the following symbols:
\begin{align}
A(\theta^t)&=\cup_yA_y(\theta^t)\quad\text{// $A_y$ is defined by Eq.\eqref{eq: def A_y}},\notag\\
m(\theta^t,N)&=\min_{\theta\in A(\theta^t)}f_N(\theta)\quad\text{// minimum in all lines $A_y$}\notag\\
&=\min_y m_y(\theta^t,N)\label{eq: def m},\\
Y_\text{min}&=\{y|m_y(\theta^t,N)=m(\theta,N)\},\notag\\
\arg m_y(\theta^t,N)&=\{\theta|\theta\in A_y(\theta^t),f_N(\theta)=m_y(\theta^t,N)\},\notag\\
\arg m(\theta^t,N)&=\{\theta|\theta\in A(\theta^t),f_N(\theta)=m(\theta^t,N)\},\notag\\
&=\cup_{y\in Y_\text{min}}\arg m_y(\theta^t,N).\label{eq: argm}
\end{align}
Then, Algorithm \ref{alg: learn} iteratively selects $\theta^{t+1}$ from $\arg m(\theta^t,N)$.

We first consider the case where the cost function is a continuous function of $\theta$ with no regularization term.
Then, the following theorem holds \citep{beck2015convergence}.
\begin{theorem}\label{thm: accumulation point}
Let $f:\mathbb R^{|X|}\to\mathbb R$ be a continuous cost function such that $B=\{\theta|f(\theta)\le f(\theta^0)\}$ is a bounded close set.
Then, any accumulation point of $\{\theta^t\}(t\in[0,\infty))$ is an axis minimum of $f$.
\end{theorem}
The proof is provided in Appendix.
Here, if $f(\theta)$ has an unique axis minimum at $\theta=a$, the following corollary is derived from Theorem \ref{thm: accumulation point}.
\begin{corollary}\label{cor: continuous convergence}
Let $f$ be a function satisfying the conition in Theorem \ref{thm: accumulation point}.
If $f$ has a unique axis minimum at $\theta=\theta_\text{min}$, then $\theta_\text{min}$ is also the global minimum and $\lim_{t\to\infty}\theta^t=\theta_\text{min}$.
\end{corollary}
The proof is provided in Appendix.

Let $q$ be a positive distribution.
By Corollary \ref{cor: continuous convergence}, in the case where the cost function is $f(\theta)=KL(q\|p_\theta)$, where $q$ is a positive distribution\footnote{\normalsize Positivity is needed to keep $B$ bounded.}, the equation $\lim_{t\to\infty}KL(q\|p_{\theta^t})=0$ holds.

Then, we extend the cost function to $f_N(\theta)=f(\theta)+r(\theta,N)$, where $f:\mathbb R^{|X|}\to\mathbb R$ is a continuous function having the unique global minimum at $\theta=\theta_\text{min}$, and $r(\theta,N)$ be a regularization term in Eq.\eqref{eq: cost}.
The following theorem holds.
\begin{theorem}\label{thm: main theorem}
Let $f$ be a function satisfying the conition in Theorem \ref{thm: accumulation point}.
Then,
$$\lim_{N\to\infty}\lim_{t\to\infty}f(\theta^t(N))=\min_\theta f(\theta)$$
holds.
\end{theorem}
Here, do not confuse $\lim_{N\to\infty}\lim_{t\to\infty}f(\theta^t(N))$ with $\lim_{t\to\infty}\lim_{N\to\infty}f(\theta^t(N))$.
$$\lim_{t\to\infty}\lim_{N\to\infty}f(\theta^t(N))=\min_\theta f(\theta)$$
is trivial by Corollary \ref{cor: continuous convergence}, while
$$\lim_{N\to\infty}\lim_{t\to\infty}f(\theta^t(N))=\min_\theta f(\theta)$$
needs a proof.
The proof is provided in Appendix.
In the case where $p_d$ is a positive distribution for sufficiently large $N$ and $f(\theta)=KL(p_d|p_\theta)$, Theorem \ref{thm: main theorem} guarantees:
$$\lim_{N\to\infty}\lim_{t\to\infty}KL(p_d\|p_{\theta^t(N)})=0.$$

\section{Experiments}
In this section, to demonstrate the performance of the FSLL model, we compare a full-span log-linear model that we refer to as {\it FL} with two Boltzmann machines that we refer to as {\it BM-DI} and {\it BM-PCD}.

\subsection{Full-Span Log-Linear Model FL}
FL is a full-span log-linear model that has been described in Section \ref{sec: fsll}.
The model distribution of FL is given by Eq.\eqref{eq: fsllmodel}.
The cost function is given by Eq.\eqref{eq: cost}.
The learning algorithm is Algorithm \ref{alg: learn} described in Section \ref{sec: learning algorithm}.
The threshold to finish the cost minimization is determined as $\epsilon=10^{-4}$.

\subsection{Boltzmann Machine BM-DI}
BM-DI(Boltzmann machine with direct integration) is a fully connected Boltzmann machine having no hidden variables and no temperature parameter.
To examine the ideal performance of the Boltzmann machine, we do not use the Monte Carlo approximation in BM-DI to evaluate Eq.\eqref{eq: gradBM}.
The model distribution of BM-DI is given by Eq.\eqref{eq: BM}.
The cost function of BM-DI is $ KL (p_d\|p_\theta)$ and has no regularization term.

To minimize the cost function with less evaluations of the gradient vector, we used a pseudo-Newton method called {\it Broyden-Fletcher-Goldfarb-Shanno(BFGS)} algorithm\citep[Chapter 6]{nocedal2006numerical} implemented in Java Statistical Analysis Tool(JSAT) \citep{JMLR:v18:16-131}.
\subsection{Boltzmann Machine BM-PCD}
BM-PCD(Boltzman Machine with persistent contrastive divergence method) is similar to BM-DI; however, BM-DI uses {\it persistent contrastive divergence} method\citep{tieleman2008training} that is a popular Monte Carlo method in Boltzmann machine learning.
BM-PCD has some hyperparameters.
We tested various combinations of these hyperparameters and determined them as learning rate=0.01, number of Markov chains=100, length of Markov chains=10000.

\subsection{Training Data}
We prepared six training datasets.
These datasets are artificial; therefore, their true distributions $p_*(X)$ are known.
Each dataset is an independent and identically distributed (i.i.d.) dataset drawn from its true distribution.
\begin{description}
\item[Ising5x4S, Ising5x4L]These datasets were drawn from the distribution represented by the following 2D Ising model \citep[Chapter 1]{newman1999monte} with $5\times4$ nodes (Figure \ref{fig: ising5x4}).

\begin{figure}[h]
\centering\includegraphics{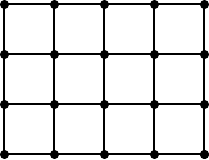}
\caption{Graphical structure of Ising5x4S/L}\label{fig: ising5x4}
\end{figure}

Every $X_i$ takes the value 0 or 1.
Ising5x4S has 1000 samples, while Ising5x4L has 100,000 samples.
The true distribution is represented as follows:
\begin{multline*}
p_*(x)=Z^{-1}\exp\left(\frac12\sum_{<i,j>}s_is_j\right),\quad
Z=\sum_x\exp\left(\frac12\sum_{<i,j>}s_is_j\right),\\
s_i=\begin{cases}
1&x_i=1\\
-1&x_i=0
\end{cases},
\end{multline*}
where $<i,j>$ denotes the adjacent variables in Figure \ref{fig: ising5x4}.
Boltzmann machines can represent $p_*$.
\item[BN20-37S, BN20-37L]These datasets were drawn from the distribution represented by the following Bayesian network with 20 nodes and 37 edges (Figure \ref{fig: BN20-37}).

\begin{figure}[h]
\centering\includegraphics{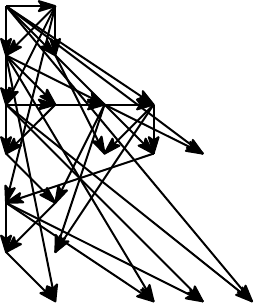}
\caption{Graphical structure of BN20-37S/L}\label{fig: BN20-37}
\end{figure}

$X_0$ has no parents, $X_1$ has $X_0$ as a parent, and the other $X_i(i\ge2)$ individually has two parents $Y_{i0},Y_{i1}\in\{X_0,...,X_{i-1}\}$.
The graphical structure and contents of the conditional distribution table of each node are determined randomly.
Every $X_i$ takes the value 0 or 1.
BN20-37S has 1000 samples, while BN20-37L has 100,000 samples.
The true distribution is represented as follows:
\begin{align*}
p_*(x)&=\prod_i p_*(x_i|y_{i0},y_{i1})\\
&=\exp\left(\sum_i\ln p_*(x_i|y_{i0},y_{i1})\right).
\end{align*}
Since $\ln p_*(x_i|y_{i0},y_{i1})$ are third-order terms, third-order Boltzmann machines can represent $p_*$.
\item[BN20-54S, BN20-54L]
These datasets were drawn from the distribution represented by the following Bayesian network with 20 nodes and 54 edges.
$X_0$ has no parents, $X_1$ has $X_0$ as a parent, $X_2$ has $X_0,X_1$ as parents, and the other $X_i(i\ge3)$ individually has three parents $Y_{i0},Y_{i1},Y_{i2}\in\{X_0,...,X_{i-1}\}$.
The graphical structure and contents of the conditional distribution table of each node are determined randomly.
Every $X_i$ takes the value 0 or 1.
BN20-54S has 1000 samples, and BN20-54L has 100000 samples.
The true distribution is represented as follows:
\begin{align*}
p_*(x)&=\prod_i p_*(x_i|y_{i0},y_{i1},y_{i2})\\
&=\exp\left(\sum_i\ln p_*(x_i|y_{i0},y_{i1},y_{i2})\right).
\end{align*}
Since $\ln p_*(x_i|y_{i0},y_{i1},y_{i2})$ are fourth-order terms, fourth-order Boltzmann machines can represent $p_*$.
\end{description}

\subsection{Experimental Platform}
All experiments were conducted on a laptop PC (CPU: Intel Core i7-6700K @4GHz; memory: 64 GB; operating system: Windows 10 Pro).
All programs were written in and executed on Java 8.

\subsection{Results}
\begin{table}[h]
\centering
\caption{Performance comparison between FL and BM}\label{table: FL and BM}
\begin{tabular}{|c|c|c|c|c|c|} \hline
Data	&Model	&$KL(p_d\|p_\theta)$	&$KL(p_*\|p_\theta)$	&\#Basis	&Time\\ \hline
Ising5x4S	&FL	&2.501nat	&0.012nat	&31	&5sec\\
&BM-DI	&2.424	&0.087	&210	&13\\
&BM-PCD	&2.504	&0.094	&210	&3\\\hline
Ising5x4L	&FL	&0.476	&0.004	&37	&9\\
&BM-DI	&0.473	&0.002	&210	&12\\
&BM-PCD	&0.528	&0.053	&210	&3\\\hline
BN20-37S	&FL	&4.355	&0.317	&39	&5\\
&BM-DI	&4.746	&0.863	&210	&17\\
&BM-PCD	&4.803	&0.903	&210	&3\\\hline
BN20-37L	&FL	&0.697	&0.026	&105	&12\\
&BM-DI	&1.422	&0.750	&210	&19\\
&BM-PCD	&1.477	&0.806	&210	&3\\\hline
BN20-54S	&FL	&3.288	&0.697	&41	&5\\
&BM-DI	&3.743	&1.301	&210	&23\\
&BM-PCD	&3.826	&1.338	&210	&3\\\hline
BN20-54L	&FL	&0.430	&0.057	&192	&23\\
&BM-DI	&1.545	&1.166	&210	&21\\
&BM-PCD	&1.620	&1.242	&210	&3\\\hline
\end{tabular}\label{table: KL}\\
$p_d$: empirical distribution of training data\quad
$p_*$: true distribution\\
\#Basis: number of used($\theta_y\ne0$) basis functions\\
Time: CPU time for learning(median of three trials)\\
\end{table}

\begin{figure}[h]
\centering\includegraphics{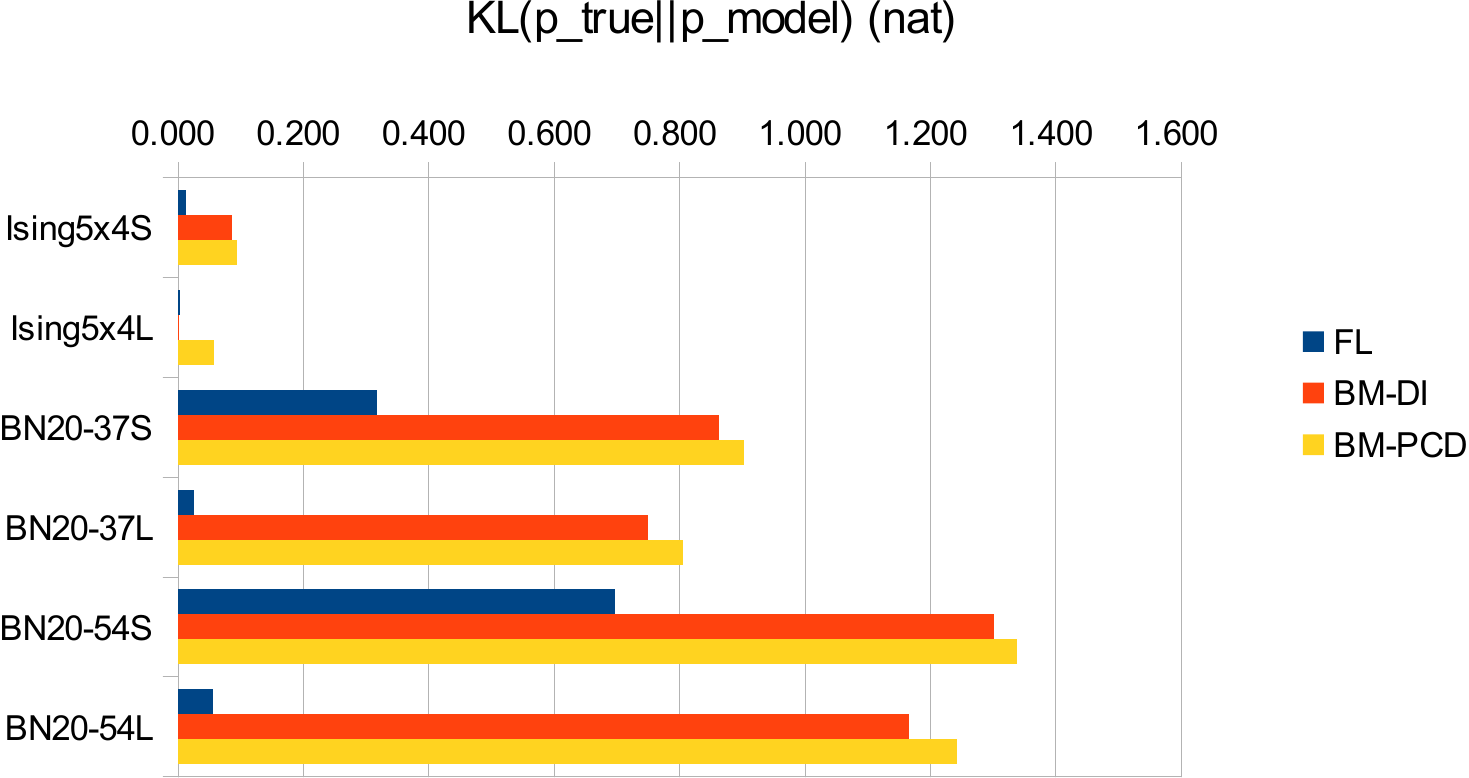}
\caption{Comparing accuracy of model by $KL(p_*\|p_\theta)$ (lower is better)}\label{fig: KL}
\end{figure}

Table \ref{table: FL and BM} represents performance comparisons between FL BM-DI and BM-PCD.
We evaluated the accuracy of the learned distribution by $ KL (p_*\|p_\theta)$.
Figure \ref{fig: KL} illustrates the comparison of $KL(p_*\|p_\theta)$.

For Ising5x4S/L, a performance difference between FL and BMs (BM-DI and BM-PCD) was not remarkable because both FL and BMs could represent the true distribution $p_*$.
The fact that $KL(p_d\|p_\theta)\gg KL(p_*\|p_\theta)$ implies that overfitting to $p_d$ was successfully suppressed.
FL used fewer basis functions than BMs used, which implies that some basis functions of BM were useless to represent $p_*$.
Regarding the accuracy of the model distribution, BM-PCD has less accuracy than FL and BM-DI have.
This disadvantage becomes noticeable when the model distribution is close to the true distribution.
Even large training data are given, some error remains in the model distribution of BM-PCD(for example, Ising5x4L).

For BN20-37S/L and BN20-54S/L, FL outperformed BMs because only FL could represent $p_*$.
To fit $p_*$, FL adaptively selected 39 basis functions for BN20-37S and 105 basis functions for BN20-37L from $|X|-1=2^{20}-1$ basis functions.
This fact implies that FL constructed a more complex model to fit $p_*$ as the training data increased.
Furthermore, a comparison of $ KL (p_*\|p_\theta)$ revealed that the accuracy of the model distribution was remarkably improved as the size of training data increased in FL.
In contrast, BMs could not fit $p_*$ even if a large training dataset was supplied.

\begin{figure}[h]
\centering\includegraphics{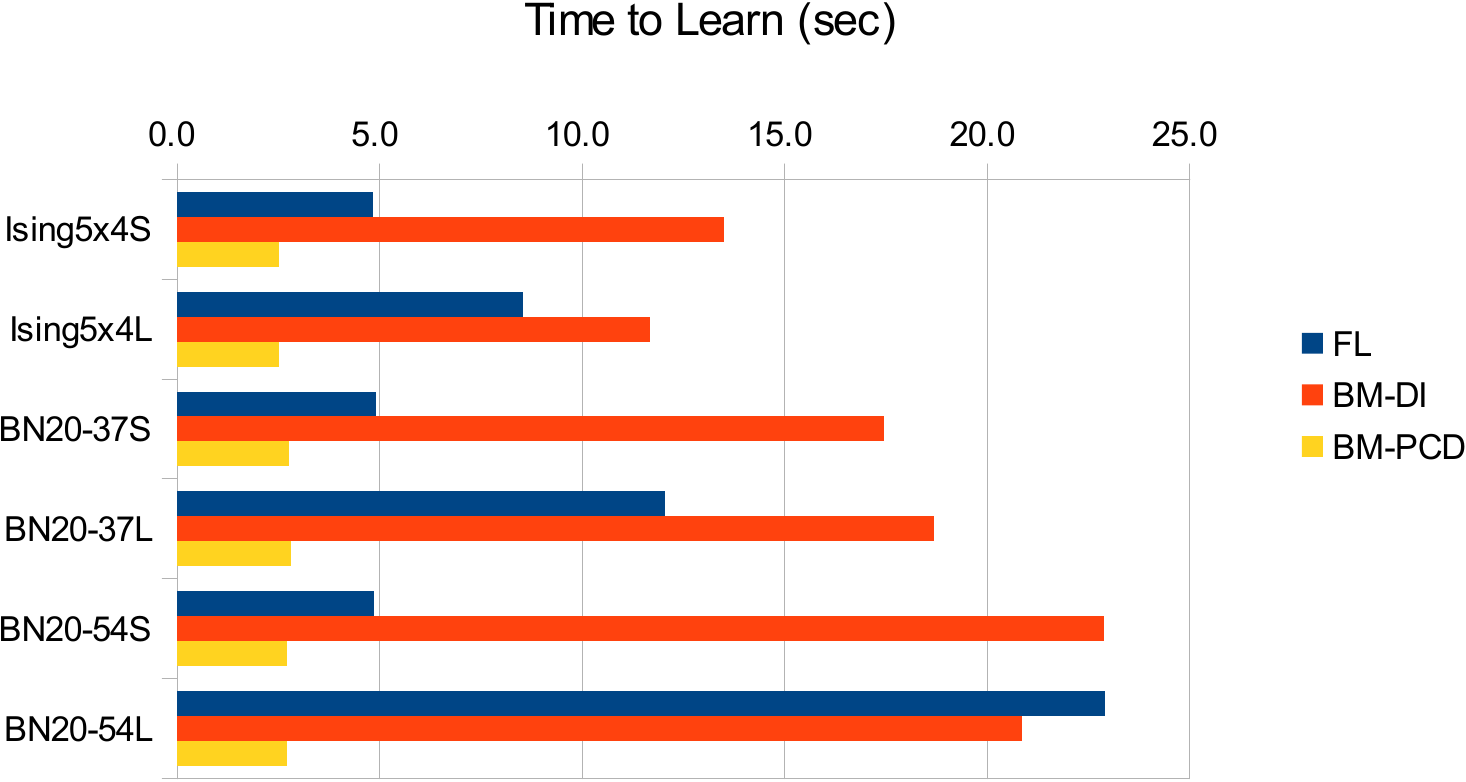}
\caption{Comparing CPU time to learn(lower is better)}\label{fig: time}
\end{figure}

Figure \ref{fig: time} illustrates the CPU time to learn the training datasets.
BM-PCD was the fastest, and FL was faster than BM-DI for five out of six training datasets.
The learning time of BM-PCD is constant because we used a fixed length(10000) of Markov chains.
FL had $|X|-1=2^{20}-1$ basis functions, while BM-DI had $n(n+1)/2=210$ basis functions.
Nevertheless, FL was faster than BM-DI.

For BN20-54L, FL takes a long time to learn because it uses 192 basis functions to construct the model distribution.
Using 192 bases functions, FL successfully constructed a model distribution that fitted $p_*$, while BMs failed.

\section{Discussion}
The major disadvantage of the FSLL model is that the FSLL model is not feasible for large problems due to memory consumption and learning speed.
If we use a typical present personal computer, the problem size should be limited as $|X|\stackrel<\sim2^{25}$.
However, as far as we use the FSLL model in this problem domain, the FSLL model is a practical model that has the following theoretical and practical advantages.

The first advantage is that the FSLL model can represent arbitrary distributions of $X$.
Furthermore, it is guaranteed that the model distribution converges to any target distribution at the limit of the training data size is infinity.

Here, let us view learning machines from an information geometry perspective.
Let $\mathcal P$ be the space of positive distributions of $X$ that can take $|X|$ values.
Then, the dimension of $\mathcal P$ is $|X|-1$, and a learning machine having $M$ parameters spans a $M$-dimensional manifold in $\mathcal P$ to represent its model distribution(we refer to this manifold as the {\it model manifold}).

Any learning machine having $M<|X|-1$ parameters cannot represent arbitrary distributions in $\mathcal P$.
Moreover, if $M<|X|-1$, there is no guarantee that the true distribution is close to the model manifold, and if the model manifold is remote from the true distribution, the machine's performance will be poor.
This poor performance is not improved even infinite training data are given.

The FSLL model extends the manifold's dimension to $|X|-1$ by introducing higher-order factors.
The model manifold becomes $\mathcal P$ itself; thus, there is no more expansion; therefore, we refer to the model as the {\it Full-Span log-linear model}.
The key of this paper is that as far as the problem size is $|X|\stackrel<\sim2^{25}$, the FSLL model becomes a feasible and practical model.

For example, suppose that we construct a full-span model by adding hidden nodes into a Boltzmann machine having 20 visible nodes.
The number of parameters of the Boltzmann machine is $E+n$, where $E$ is the number of edges and $n$ is the number of nodes.
Therefore, it is not practical to construct the full-span model for 20 visible nodes because it requires $\approx 2^{20}$ edges.

The second advantage is that the FSLL model has no hyperparameters; therefore, no hyperparameter tuning is needed.
For example, if we use a Boltzmann machine with hidden nodes that learns the true distribution with contrastive divergence methods, we need to determine hyperparameters such as the learning rate, mini-batch size, and the number of hidden, and the graphical structure of nodes.
On the other hand, the FSLL model automatically learns the training data without human participation.

\section{Conclusion and Extension}
Suppose that we let the FSLL model learn training data consisting of 20 binary variables.
The dimension of the function space spanned by possible positive distributions is $2^{20}-1$.
The FSLL model has $2^{20}-1$ parameters and can fit arbitrary positive distributions.
The FSLL model has the basis functions that have the following properties:
\begin{itemize}
\item Each basis function is a product of univariate functions.
\item The basis functions take values 1 or $-1$.
\end{itemize}
The proposed learning algorithm exploited these properties and realized fast learning.

Our experiments demonstrated the following:
\begin{itemize}
\item The FSLL model could learn the training data with 20 binary variables within 1 minute with a laptop pc.
\item The FSLL model successfully learned the true distribution underlying the training data even higher-order terms that depend on three or more variables existed.
\item The FSLL model constructed a more complex model to fit the true distribution as the training data increased; however, the learning time became longer.
\end{itemize}

In this paper, we have presented a basic version of the FSLL model; however, we can extend it as follows \citep{Takabatake14,Takabatake15}:
\begin{itemize}
\item Introducing $L_1$ regularization \citep{andrew2007scalable},
\item Introducing hidden variables.
\end{itemize}

\subsection*{Acknowledgments}
This research is supported by KAKENHI 17H01793.

\section*{Appendix}
\subsection*{Proof of Theorem \ref{thm: independence}}
For brevity and clarity of expression, we use predicate logic notation here.
The statement ``$\{f_i(X_0)g_j(X_1)\}(i\in I, j\in J)$ are linearly independent functions of $(X_0,X_1)$.'' is equivalent to the following proposition:
$$\forall x_0\forall x_1\left[\sum_{i,j}
a_{ij}f_i(x_0)g_j(x_1)=0\right]\Rightarrow\forall i\forall j
\left[a_{ij}=0\right].$$
This proposition is proved as follows:
\begin{align*}
&\forall x_0\forall x_1\left[\sum_{i,j}a_{ij}f_i(x_0)g_j(x_1)=0\right]\\
&\Rightarrow\forall x_0
\left[\forall x_1\left[\sum_j\left(\sum_i a_{ij}f_i(x_0)\right)g_j(x_1)=0\right]\right]\\
&\Rightarrow\forall x_0
\left[\forall j\left[\sum_i a_{ij}f_i(x_0)=0\right]\right]\quad
\because\{g_j\}\text{ are linearly independent}\\
&\Rightarrow\forall i\forall j\left[a_{ij}=0\right].\quad
\because\{f_i\}\text{ are linearly independent}\\
\end{align*}
\rightline{Q.E.D.}

\subsection*{Derivation of Eq.\eqref{eq: barTheta}}
\begin{align*}
\bar\theta_y(\theta)&=\partial_y\ln Z\\
&=\frac1Z\partial_y Z\\
&=\frac1Z\sum_x\partial_y e^{l_\theta(x)}\\
&=\frac1Z\sum_x\Phi_y(x)e^{l_\theta(x)}\\
&=\sum_x p_\theta(x)\Phi_y(x)\\
&=\left<\Phi_y\right>_{p_\theta}
\end{align*}

\subsection*{Derivation of Eq.\eqref{eq: partial_y bar_y}}
\begin{align*}
\partial_y\bar\theta_y&=\partial_y\left<\Phi_y\right>_{p_\theta}\\
&=\sum_x\Phi_y(x)\partial_y p_\theta(x)\\
&=\sum_x\Phi_y(x)p_\theta(x)\partial_y\ln p_\theta(x)\\
&=\sum_x\Phi_y(x)p_\theta(x)\partial_y(l_\theta(x)-\ln Z)\\
&=\sum_x\Phi_y(x)p_\theta(x)(\Phi_y(x)-\partial_y\ln Z)\\
&=\sum_x p_\theta(x)\Phi_y(x)^2-\bar\theta_y\sum_x p_\theta(x)\Phi_y(x)\\
&=\left<\Phi_y^2\right>_{p_\theta}-\bar\theta_y^2
\end{align*}

\subsection*{Derivation of Eq.\eqref{eq: general solution}}
\begin{align}
1&=\frac{\partial\bar\theta_y}{1-\bar\theta_y^2}\quad
\text{(by Eq.\eqref{eq: ordinary diff})}\notag\\
&=\frac12\left(\frac{\partial_y\bar\theta_y}{1+\bar\theta_y}
+\frac{\partial_y\bar\theta_y}{1-\bar\theta_y}\right)\notag\\
&=\frac12\left(\partial_y\ln(1+\bar\theta_y)-\partial_y\ln(1-\bar\theta_y)\right)
\label{eq: to solution}
\end{align}
Integrating Eq.\eqref{eq: to solution}, we obtain
\begin{align}\label{eq: derivation of general solution}
\frac12\ln\frac{1+\bar\theta_y}{1-\bar\theta_y}&=
\int 1d\theta_y\notag\\
&=\theta_y-c,
\end{align}
where $c$ is a constant of integration.
Since the left side of Eq.\eqref{eq: derivation of general solution} equals $\tanh^{-1}(\bar\theta_y)$, we obtain the equation $\bar\theta_y=\tanh(\theta_y-c)$.

\subsection*{Derivation of Eq.\eqref{eq: gradKL}}
\begin{align*}
\partial_y KL(p_d\|p_\theta)
&=\partial_y\left(\left<\ln p_d\right>_{p_d}-\left<\ln p_\theta\right>_{p_d}\right)\\
&=-\partial_y\left<\ln p_\theta\right>_{p_d}\\
&=-\partial_y\left<l_\theta-\ln Z\right>_{p_d}\\
&=-\left<\partial_y l_\theta\right>_{p_d}+\partial_y\ln Z\\
&=-\left<\Phi_y\right>_{p_d}+\bar\theta\\
&=\bar\theta_y-\bar d
\end{align*}

\subsection*{Derivation of Eq.\eqref{eq: KL(p_d||p_theta)}}
\begin{align*}
\int_{\theta_y^0}^{\theta_y}\bar\theta_y(u)-\bar d_y du
&=\int_{\theta_y^0}^{\theta_y}\tanh(u-c)du
-\bar d_y(\theta_y-\theta_y^0)\quad
\text{(by Eq.\eqref{eq: general solution})}\\
&=\left[\ln\cosh(u-c)\right]_{u=\theta_y^0}^{\theta_y}
-\bar d_y(\tanh^{-1}\bar\theta_y-\tanh^{-1}\bar\theta_y^0)\\
&=\left[\frac12\ln\cosh^2(u-c)\right]_{u=\theta_y^0}^{\theta_y}
-\bar d_y\left(\frac12\ln\frac{1+\bar\theta_y}{1-\bar\theta_y}
-\frac12\ln\frac{1+\theta_y^0}{1-\theta_y^0}\right)\\
&=\left[\frac12\ln\frac1{1-\tanh^2(u-c)}\right]_{u=\theta_y^0}^{\theta_y}
-\frac{\bar d_y}2\ln\frac{1+\bar\theta_y}{1-\bar\theta_y}
\frac{1-\bar\theta_y^0}{1+\bar\theta_y^0}\\
&=\frac12\left(\ln\frac1{1-\bar\theta_y^2}-\ln\frac1{1-(\bar\theta_y^0)^2}\right)
-\frac{\bar d_y}2\ln\frac{1+\bar\theta_y}{1-\bar\theta_y}
\frac{1-\bar\theta_y^0}{1+\bar\theta_y^0}\\
&=\frac12\ln\frac{1+\bar\theta_y^0}{1+\bar\theta_y}
\frac{1-\bar\theta_y^0}{1-\bar\theta_y}
-\frac{\bar d_y}2\ln\frac{1+\bar\theta_y}{1-\bar\theta_y}
\frac{1-\bar\theta_y^0}{1+\bar\theta_y^0}\\
&=-\frac{1+\bar d_y}2\ln(1+\bar\theta_y)-\frac{1-\bar d_y}2\ln(1-\bar\theta_y)\\
&\quad
+\frac{1+\bar d_y}2\ln(1+\bar\theta_y^0)
+\frac{1-\bar d_y}2\ln(1-\bar\theta_y^0)\\
&=\frac{1+\bar d_y}2\ln\frac{1+\bar\theta_y^0}{1+\bar\theta_y}
+\frac{1-\bar d_y}2\ln\frac{1-\bar\theta_y^0}{1-\bar\theta_y}
\end{align*}

\subsection*{Derivation of Eq.\eqref{eq: lower bound}}
\begin{align*}
\Delta&=\frac{1+\bar d_y}2\ln\frac{1+\bar\theta^0_y}{1+\bar d_y}
+\frac{1-\bar d_y}2\ln\frac{1-\bar\theta^0_y}{1-\bar d_y}+r_y\\
&=\frac{1+\bar d_y}2\left(-\ln\frac{1+\bar d_y}{1+\bar\theta^0_y}\right)
+\frac{1-\bar d_y}2\left(-\ln\frac{1-\bar d_y}{1-\bar\theta^0_y}\right)+r_y\\
&\ge\frac{1+\bar d_y}2\left(1-\frac{1+\bar d_y}{1+\bar\theta^0_y}\right)
+\frac{1-\bar d_y}2\left(1-\frac{1-\bar d_y}{1-\bar\theta^0_y}\right)+r_y\\
&\quad
\because\frac{1+\bar d_y}2\ge0,\quad
\frac{1-\bar d_y}2\ge0,\quad
-\ln x\ge1-x\\
&=\frac{1+\bar d_y}2\frac{\bar\theta_y^0-\bar d_y}{1+\bar\theta_y^0}
-\frac{1-\bar d_y}2\frac{\bar\theta_y^0-\bar d_y}{1-\bar\theta_y^0}+r_y\\
&=\frac{\bar\theta_y^0-\bar d_y}2
\left(\frac{1+\bar d_y}{1+\bar\theta_y^0}-\frac{1-\bar d_y}{1-\bar\theta_y^0}\right)+r_y\\
&=\frac{\bar\theta_y^0-\bar d_y}2
\frac{(1+\bar d_y)(1-\bar\theta_y^0)-(1-\bar d_y)(1+\bar\theta_y^0)}
{(1+\bar\theta_y^0)(1-\bar\theta_y^0)}+r_y\\
&=\frac{\bar\theta_y^0-\bar d_y}2
\frac{2\bar d_y-2\bar\theta_y^0}{1-(\bar\theta_y^0)^2}+r_y\\
&=-\frac{(\bar\theta_y^0-\bar d_y)^2}{1-(\bar\theta_y^0)^2}+r_y\\
&=\underline\Delta
\end{align*}

\subsection*{Proof of Theorem \ref{thm: accumulation point}}
Since $B$ is a bounded closed set, $\{\theta^t\}$ has one or more accumulation point(s) in $B$.

\begin{figure}[h]
\centering\includegraphics{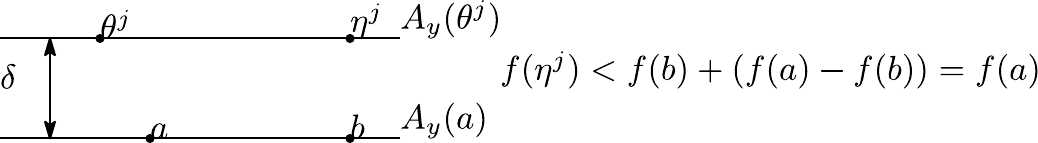}
\caption{Existence of $\theta^j$ such that $f(\eta^j)<f(a)$}\label{fig: theorem1}
\end{figure}

As an assumption of a proof by contradiction, assume that $a$ is an accumulation point of $\{\theta^t\}$ and the proposition
$$\exists b\in A_y(a),\quad f(b)<f(a)$$
holds.
Let $\eta^j\in A_y(\theta^j)$ be the point such that $\eta^j_y=b_y$.
Since $f(a)-f(b)>0$, $f(\theta)$ is continuous at $\theta=b$, and $a$ is an accumulation point of $\{\theta^t\}$, the proposition
$$\exists\theta_j,\quad f(\eta^j)<f(a)$$
holds($\because$ in Fig.\ref{fig: theorem1}, $f(\eta^j)<f(a)$ for sufficiently small $\delta$).
\begin{figure}
\end{figure}

Here, $\eta^j\in A_y(\theta^j)$ and the inequality
$$f(\theta^{j+1})=\min_{\theta\in\cup_y A_y(\theta^j)}f(\theta)\le f(\eta^j)$$
holds.
Therefore, the inequality
$$f(\theta^{j+1})\le f(\eta^j)<f(a)$$
holds.
Since $f(\theta^t)$ monotonically decreases as $t$ increases, the proposition
$$\forall t\ge j+1,\quad f(\theta^t)\le f(\eta^j)<f(a)$$
holds.
Since $f(\theta)$ is continuous at $\theta=a$, no subsequence of $\{\theta^t\}$ can converges to $a$, that is, $a$ is not an accumulation point of $\{\theta^t\}$.
This fact contradicts the assumption we made at the beginning of this proof.

\rightline{QED}

\subsection*{Proof of Corollary \ref{cor: continuous convergence}}
By Theorem \ref{thm: accumulation point}, any accumulation point of $\{\theta^t\}$ is an axis minimum of $f$, however, the axis minimum of $f$ is unique; therefore, $\theta=a$ is a unique accumulation point of $\{\theta^t\}$, and $\theta^t\to a(t\to\infty)$.

\rightline{QED}

\subsection*{Proof of Theorem \ref{thm: main theorem}}
Let us define the following symbols:
\begin{align*}
m_y(\theta^t,\infty)&=\min_{\theta\in A_y(\theta^t)}f(\theta),\\
\arg m_y(\theta^t,\infty)&=\{\theta|\theta\in A_y(\theta^t), f(\theta)=m_y(\theta^t,\infty)\}.
\end{align*}

\begin{figure}[h]
\centering\includegraphics{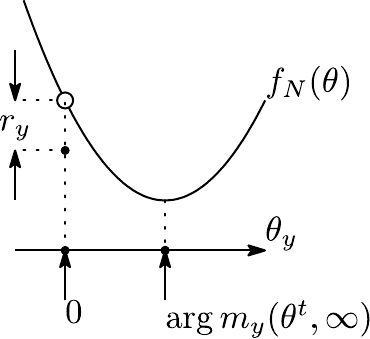}
\caption{View of $f_N(\theta$)}\label{fig: f_N}
\end{figure}

Figure \ref{fig: f_N} illustrates the sectional view of $f_N(\theta)$ along the line $A_y(\theta^t)$.
As shown in Fig.\ref{fig: f_N}, $f_N(\theta)$ has a gap with depth $r_y$(Eq.\eqref{eq: cost}) at $\theta_y=0$.
Here, we can ignore the gap at $\theta_y=0$ for sufficiently large $N$, that is, the proposition
$$\exists N'\forall N,\quad N>N'\Longrightarrow\arg m_y(\theta^t,N)=\arg m_y(\theta^t,\infty)$$
holds.
Moreover, the proposition
\begin{equation}\label{eq: mN=minf}
\exists N'\forall N\forall y,\quad N>N'\Longrightarrow\arg m_y(\theta^t,N)=\arg m_y(\theta^t,\infty)
\end{equation}
holds.
By Eq.\eqref{eq: argm}, the proposition
$$\exists N'\forall N,\quad N>N'\Longrightarrow\arg m(\theta^t,N)=\arg m(\theta^t,\infty)$$
holds.
Here, we compare a sequence $\{\theta^t(N)\}$ with the cost function $f_N$ and a sequence $\{\theta^t(\infty)\}$ with the cost function $f(=f_\infty)$.
By Corollary \ref{cor: continuous convergence}, $\lim_{t\to\infty}f(\theta^t(\infty))=\min_\theta f(\theta)$, that is,
\begin{equation}\label{eq: lim f}
\forall\epsilon>0\exists t'\forall t,\quad t\ge t'\Longrightarrow
f(\theta^t(\infty))<\min_\theta f(\theta)+\epsilon
\end{equation}
holds.
Here, let $T(N)$ be the smallest $t$ such that $m(\theta^{t+1},N)\ne m(\theta^{t+1},\infty)$.
Then,
$\lim_{N\to\infty}T(N)=\infty$, that is,
\begin{equation}\label{eq: lim T}
\forall t'\exists N,\quad T(N)\ge t'
\end{equation}
holds.
By Eq.\eqref{eq: lim f} and Eq.\eqref{eq: lim T}, the proposition
\begin{equation}\label{eq: forall t'}
\forall\epsilon>0\exists N,\forall t,\quad
t\ge T(N)\Longrightarrow f(\theta^t(\infty))<\min_\theta f(\theta)+\epsilon
\end{equation}
holds; and therefore, the proposition
\begin{equation}\label{eq: bound f inf}
\forall\epsilon>0\exists N,\quad f(\theta^{T(N)}(\infty))<\min_\theta f(\theta)+\epsilon
\end{equation}
also holds.
Here, by the definition of $T(N)$, $\theta^{T(N)}(\infty)=\theta^{T(N)}(N)$.
Therefore, we can modify Eq.\eqref{eq: bound f inf} as
\begin{equation}\label{eq: bound f N}
\forall\epsilon>0\exists N,\quad f(\theta^{T(N)}(N))<\min_\theta f(\theta)+\epsilon.
\end{equation}
Since $f(\theta^t(N))$ monotonically decreases as $t$ grows,
$$\forall\epsilon>0\exists N\forall t,\quad t\ge T(N)\Longrightarrow
f(\theta^t(N))<\min_\theta f(\theta)+\epsilon.$$
Using notation of $\lim$, we obtain
$$\lim_{N\to\infty}\lim_{t\to\infty}f(\theta^t(N))=\min_\theta f(\theta).$$

\rightline{QED}

\bibliographystyle{APA}
\bibliography{library}

\begin{thebibliography}{}

\bibitem[\protect\astroncite{Ackley et~al.}{1985}]{Ackley1985}
Ackley, D.~H., Hinton, G.~E., and Sejnowski, T.~J. (1985).
\newblock {A learning algorithm for {B}oltzmann machines}.
\newblock {\em Cognitive Science}, 9(1):147--169.

\bibitem[\protect\astroncite{Amari}{2016}]{amari2016information}
Amari, S. (2016).
\newblock {\em {Information geometry and its applications}}, volume 194.
\newblock Springer.

\bibitem[\protect\astroncite{Andrew and Gao}{2007}]{andrew2007scalable}
Andrew, G. and Gao, J. (2007).
\newblock {Scalable training of $L_1$-regularized log-linear models}.
\newblock In {\em Proceedings of the 24th international conference on Machine
  learning}, pages 33--40. ACM.

\bibitem[\protect\astroncite{Beck}{2015}]{beck2015convergence}
Beck, A. (2015).
\newblock On the convergence of alternating minimization for convex programming
  with applications to iteratively reweighted least squares and decomposition
  schemes.
\newblock {\em SIAM Journal on Optimization}, 25(1):185--209.

\bibitem[\protect\astroncite{{Dennis Jr} and
  Mor{\'{e}}}{1977}]{dennis1977quasi}
{Dennis Jr}, J.~E. and Mor{\'{e}}, J.~J. (1977).
\newblock {Quasi-Newton methods, motivation and theory}.
\newblock {\em SIAM review}, 19(1):46--89.

\bibitem[\protect\astroncite{Fino and Algazi}{1976}]{fino1976unified}
Fino, B.~J. and Algazi, V.~R. (1976).
\newblock {Unified matrix treatment of the fast Walsh-Hadamard transform}.
\newblock {\em IEEE Transactions on Computers}, 25(11):1142--1146.

\bibitem[\protect\astroncite{Newman and Barkema}{1999}]{newman1999monte}
Newman, M. and Barkema, G. (1999).
\newblock {\em {Monte carlo methods in statistical physics chapter 1-4}},
  volume~24.
\newblock Oxford University Press: New York, USA.

\bibitem[\protect\astroncite{Nocedal and Wright}{2006}]{nocedal2006numerical}
Nocedal, J. and Wright, S. (2006).
\newblock {\em {Numerical optimization}}.
\newblock Springer Science \& Business Media.

\bibitem[\protect\astroncite{Pratt et~al.}{1969}]{Pratt1969}
Pratt, W.~K., Andrews, H.~C., and Kane, J. (1969).
\newblock {Hadamard Transform Image Coding}.
\newblock {\em Proceedings of the IEEE}, 57(1):58--68.

\bibitem[\protect\astroncite{Raff}{2017}]{JMLR:v18:16-131}
Raff, E. (2017).
\newblock {JSAT: Java Statistical Analysis Tool, a Library for Machine
  Learning}.
\newblock {\em Journal of Machine Learning Research}, 18(23):1--5.

\bibitem[\protect\astroncite{Rissanen}{2007}]{Rissanen07}
Rissanen, J. (2007).
\newblock {\em {Information and Complexity in Statistical Modeling}}.
\newblock Springer.

\bibitem[\protect\astroncite{Sejnowski}{1986}]{Sejnowski1986}
Sejnowski, T.~J. (1986).
\newblock {Higher-order Boltzmann machines}.
\newblock In {\em AIP Conference Proceedings}, volume 151, pages 398--403.

\bibitem[\protect\astroncite{Smith}{2010}]{Smith2010}
Smith, W.~W. (2010).
\newblock {\em {Handbook of Real-Time Fast Fourier Transforms}}.
\newblock IEEE New York.

\bibitem[\protect\astroncite{Takabatake and Akaho}{2014}]{Takabatake14}
Takabatake, K. and Akaho, S. (2014).
\newblock {Basis Functions for Fast Learning of Log-linear Models (in
  Japanese)}.
\newblock {\em IEICE technical report}, 114(306):307--312.

\bibitem[\protect\astroncite{Takabatake and Akaho}{2015}]{Takabatake15}
Takabatake, K. and Akaho, S. (2015).
\newblock {Full-span log-linear model with $L_1$ regularization and its
  performance (in Japanese)}.
\newblock {\em IEICE technical report}, 115(323):153--157.

\bibitem[\protect\astroncite{Tieleman}{2008}]{tieleman2008training}
Tieleman, T. (2008).
\newblock Training restricted boltzmann machines using approximations to the
  likelihood gradient.
\newblock In {\em Proceedings of the 25th international conference on Machine
  learning}, pages 1064--1071.

\end{thebibliography}
\end{document}